\documentclass{llncs}
\usepackage[draft]{todonotes} 
\usepackage[linesnumbered,ruled]{algorithm2e}
\usepackage{amssymb}
\usepackage{bm}
\usepackage{amsfonts}
\usepackage{subcaption}
\usepackage[cmex10]{amsmath}
\usepackage[T1]{fontenc}
\usepackage{mathtools}    

\usepackage{bm}
\usepackage{mathtools}
\usepackage[linesnumbered,ruled]{algorithm2e}
\usepackage{setspace}
\usepackage{siunitx}
\usepackage{hyperref}
\usepackage{graphicx}
\usepackage{float}
\usepackage{subcaption}
\usepackage[utf8]{inputenc}
\usepackage[english]{babel}
\usepackage{authblk}
\usepackage{colortbl}
\usepackage{ctable}
\newcolumntype{+}{>{\global\let\currentrowstyle\relax}}
\newcolumntype{^}{>{\currentrowstyle}}

\newcommand{\PreserveBackslash}[1]{\let\temp=\\#1\let\\=\temp}
\newcolumntype{C}[1]{>{\PreserveBackslash\centering}p{#1}}
\newcolumntype{R}[1]{>{\PreserveBackslash\raggedleft}p{#1}}
\newcolumntype{L}[1]{>{\PreserveBackslash\raggedright}p{#1}}

\def\tablename{Table}

\usepackage{multirow}
\usepackage{hyperref}
\usepackage{flushend}

\usepackage{soul}

\begin{document}

\title{The Deep Poincar\'e Map: A Novel Approach for Left Ventricle Segmentation}
\author{Yuanhan Mo\inst{1} \and
Fangde Liu\inst{1}\and
Douglas McIlwraith\inst{1} \and 
Guang Yang\inst{2} \and 
Jingqing Zhang\inst{1} \and 
Taigang He\inst{3} \and 
Yike Guo\inst{1}}

\institute{Data Science Institute, Imperial College London \and
National Heart \& Lung Institute, Imperial College London
\and
St George's Hospital, University of London}

\maketitle
\begin{abstract}
\noindent Precise segmentation of the left ventricle (LV) within cardiac MRI images is a prerequisite for the quantitative measurement of heart function. However, this task is challenging due to the limited availability of labeled data and motion artifacts from cardiac imaging. In this work, we present an iterative segmentation algorithm for LV delineation. By coupling deep learning with a novel dynamic-based labeling scheme, we present a new methodology where a policy model is learned to guide an agent to travel over the the image, tracing out a boundary of the ROI -- using the magnitude difference of the Poincar\'e map as a stopping criterion. Our method is evaluated on two datasets, namely the Sunnybrook Cardiac Dataset (SCD) and data from the STACOM 2011 LV segmentation challenge. Our method outperforms the previous research over many metrics. In order to demonstrate the transferability of our method we present encouraging results over the STACOM 2011 data, when using a model trained on the SCD dataset.
\end{abstract}

\section{Introduction}
Automatic left ventricle (LV) segmentation from cardiac MRI images is a prerequisite to quantitatively measure cardiac output and perform functional analysis of the heart. However, this task is still challenging due to the requirement for relatively large manually delineated datasets when using statistical shape models or (multi-)atlas based methods. Moreover, as the heart and chest are constantly in motion the resulting images may contain motion artifacts with low signal to noise ratio. Such poor quality images can further complicate the subsequent LV segmentation.   

Deep learning based methods have been proved effective for LV segmentation \cite{avendi2016combined,tan2017convolutional,ngo2017combining}. A detailed survey of the state-of-the-art lies outside the scope of this paper, but can be found elsewhere \cite{xue2018full}. Such approaches are often based on, or extend image recognition research, and thus require large training datasets that are not always available for the cardiac MRI. To the best of our knowledge, there is very limited work using significant prior information to reduce the amount of training data required while maintaining a robust performance for LV segmentation.\par
In this paper, we propose a novel LV segmentation method called the Deep Poincar\'e Map (DPM). Our DPM method encapsulates prior information with a dynamical system employed for labeling. Deep learning is then used to learn a displacement policy for traversal around the region of interest (ROI). Given an image, a CNN-based policy model can navigate an agent over the cardiac MRI image, moving toward a path which outlines the LV. At each time step, a next step policy (a 2D displacement) is given by our trained policy model, taking into account the surrounding pixels in a local squared patch. In order to learn the displacement policy, the DPM requires a data transformation step which converts the labeled images into a customized dynamic capturing the prior information around the ROI. An important property of DPM is that no matter where the agent starts, it will finally travel around the ROI. This behavior is guaranteed by the existence of a limit cycle using our customized dynamic.\par
The main contributions of this work are as follows. 
(1) The DPM integrates prior information in the form of the context of the image surrounding the ROI. It does this by combining a dynamical system with a deep learning method for building a displacement policy model, and thus requires much less data that traditional deep learning methods.
(2) The DPM is rotationally invariant. Because our next step policy predictor is trained with locally oriented patches, the orientation of the image with respect to the ROI is irrelevant.
(3) The DPM is strongly transferable. Because the context of the segmentation boundary is considered, our method generalizes well to previously unseen images with the same or similar contexts.

\begin{figure}[h!]
\begin{center}
  \includegraphics[scale=0.22]{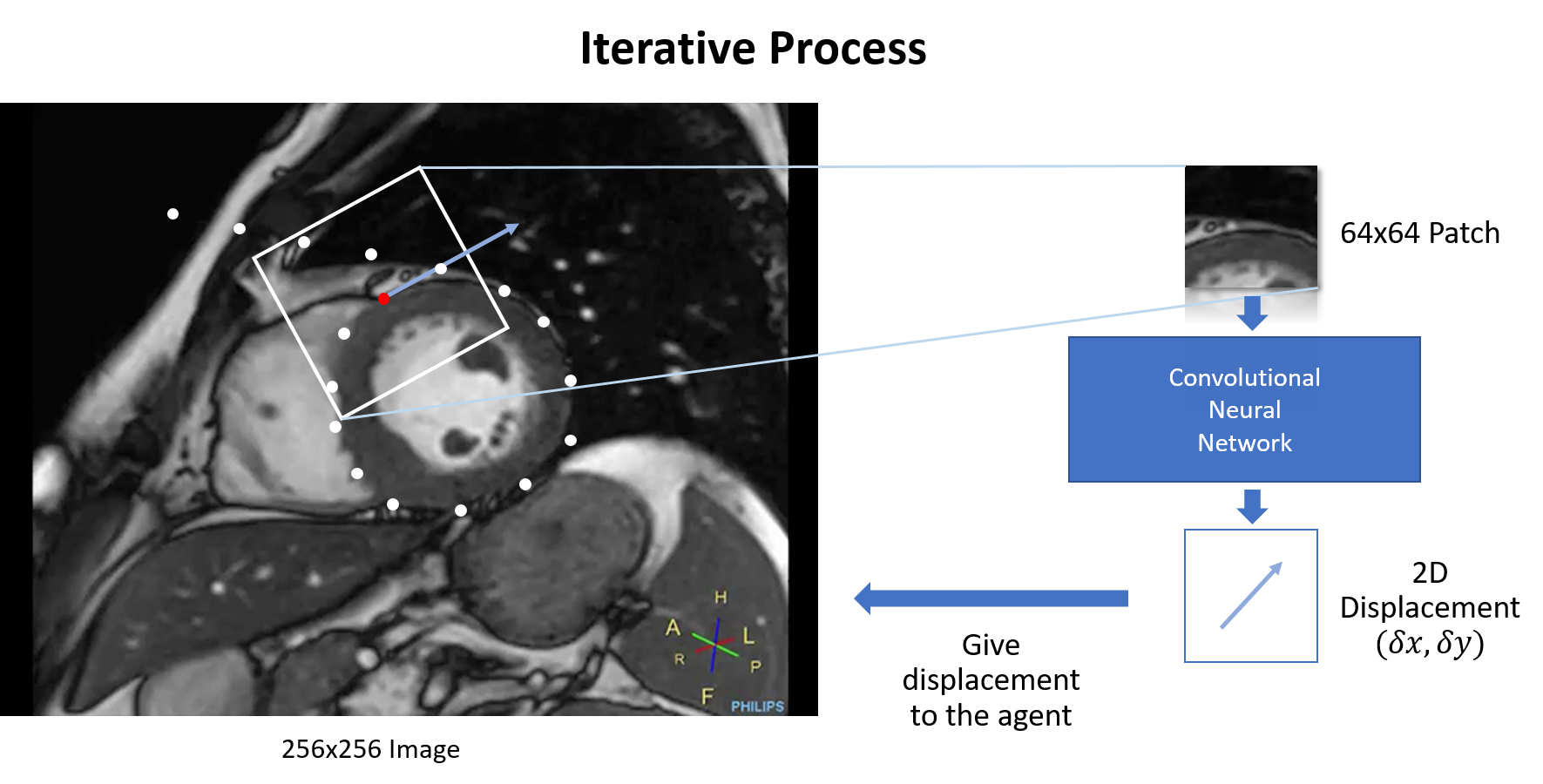}
  \caption{The red dot denotes the current position of the agent. In each time step, the DPM extracts a locally oriented patch from the original image. The extracted patch will be fed into a CNN to predict the next step displacement for the agent. After a finite number of iterations, A trajectory will be created by the agent. The magnitude of the Poincar\'e map is used to determine the final periodic orbit which is coincident with the boundary around the ROI.}
  \label{wholeprocess_lite_1}
  \end{center}
\vspace{-1.5em}  
\end{figure}

\section{Methodology}
\label{sec:method}
As shown in Fig~\ref{wholeprocess_lite_1}, 
The DPM uses a CNN-based policy model, trained on locally oriented patches from manually segmented data, to navigate an agent over a cardiac MRI image (256x256) using a locally oriented square patch (64x64) as its input. The agent creates a trajectory over the image tracing the boundary of the LV -- no matter where the agent starts on the image. A crucial prerequisite of this methodology is the creation of a vector field whose limit cycle is equal to the boundary surrounding the ROI. This can be seen in Fig~\ref{poincaresection_b}. In the following sections we will discuss the DPM methodology in detail, namely 1) the creation of a customized dynamic (i.e. a vector field) with a limit cycle around the ROI of the manually delineated images. 2) The creation of a patch-policy predictor. 3) The stopping criterion using the Poincar\'e map.\par

\subsection{Generating a Customized Dynamic}
A typical training dataset for segmentation consists of many image-to-label pairs. A label is a binary map that has the same resolution as its corresponding image. In each label, pixels of ground truth will be set to 1 while the background will be set to 0. Conversely, in our system, we firstly construct a customized dynamic (a vector field) for each labeled training instance. The constructed dynamic results in a unique limit cycle which is placed exactly on the boundary of the ROI. \par

\begin{figure}[h!]
\begin{center}
  \includegraphics[scale=0.357]{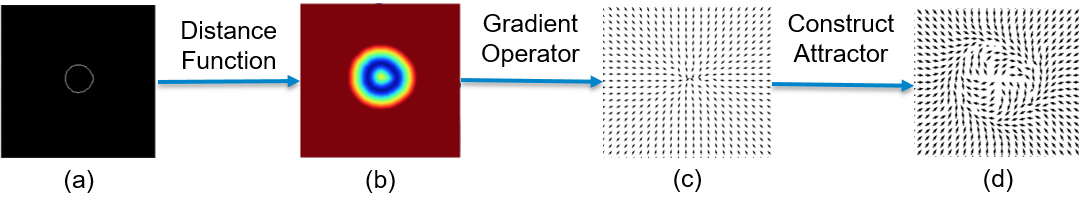}
  \caption{Demonstrating customized dynamic creation from label data.}
  \label{label_to_dynamic}
  \end{center}
\vspace{-1.5em}  
\end{figure}

\begin{figure}[h!]
\label{figtwo}
  \begin{subfigure}[b]{0.43\textwidth}
    \includegraphics[width=\textwidth]{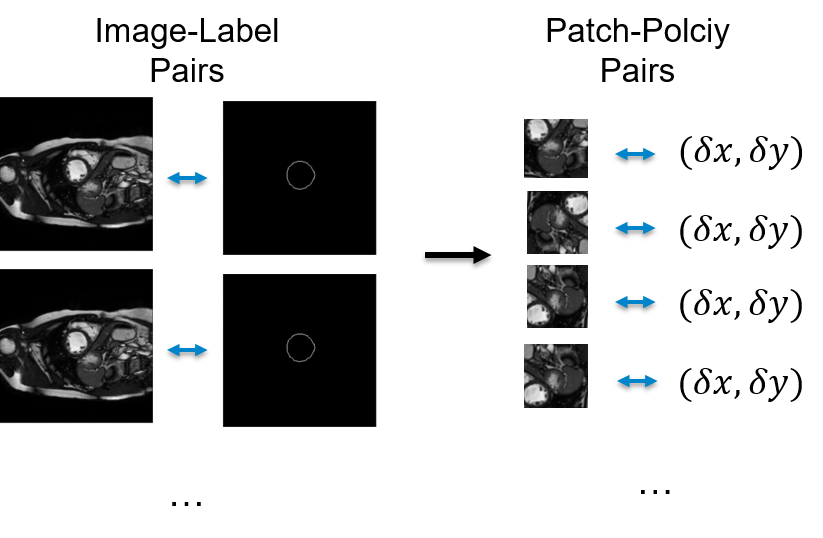}
    \caption{}
    \label{pplaims}
  \end{subfigure}
  \hfill
  \begin{subfigure}[b]{0.43\textwidth}
    \includegraphics[width=\textwidth]{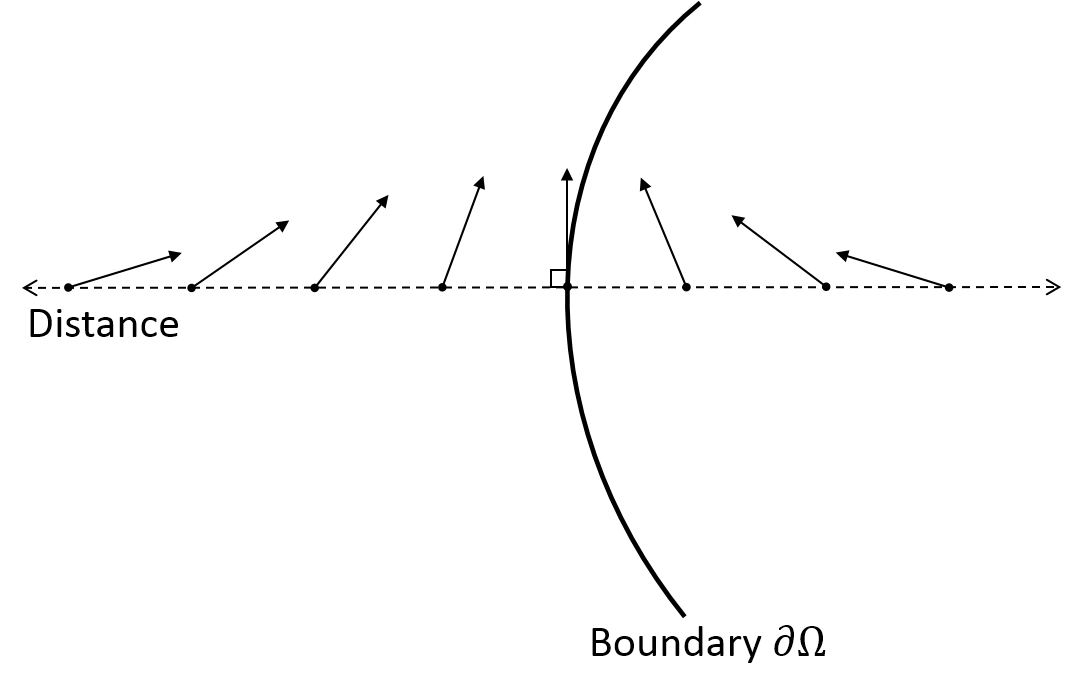}
    \caption{}
    \label{intersec}
  \end{subfigure}
  \caption{(a) Transferring original dataset to patch-policy pairs. Patch-policy pairs are the training data for policy CNN. (b) The distance between a pixel and the boundary determines how much a vector will be rotated.}
\vspace{-1.5em}  
\end{figure}

To illustrate, let us consider an example indicated in Fig~\ref{label_to_dynamic}. Consider a label of a training instance as a continuous 2D space $\mathbb{R}^2$ (a label with theoretical infinite resolution), we define the ground truth contour as a subspace $\Omega \subseteq \mathbb{R}^2$ as shown in step (a) in Fig~\ref{label_to_dynamic}. To construct a dynamic in $\mathbb{R}^2$ where a limit cycle exists and is exactly the boundary $\partial{\Omega}$, we firstly introduce the distance function $S(p)$:
\begin{equation}
S(p) =\left\{
                \begin{array}{ll}
                  d(p, \partial{\Omega}) \quad 	& \textrm{if } p \textrm{ is not on } \partial{\Omega} \\
                  0      \quad & \textrm{if } p  \textrm{ is on }  \partial{\Omega}
                \end{array}
              \right.
\label{distancefunction}
\end{equation}
$d(p, \partial{\Omega})$ denotes the infimum Euclidean distance from $p$ to the boundary $\partial{\Omega}$. Eqt~\ref{distancefunction} is used to create a scalar field from a binary image as shown in step (b) in Fig~\ref{label_to_dynamic}. In order to build the customized dynamic, we need to create a vector field from this scalar field. A gradient operator is applied to create dynamic equivalent to the active contour \cite{CootesActiveModels} as shown in step (c) in Fig~\ref{label_to_dynamic}.
This gradient operator is expressed as Eqt~\ref{ac}.
\begin{equation}
\label{ac}
\frac{dp}{dt} =  \nabla_p{S(p)}, 
\end{equation}
Our final step adds a limit cycle onto the system by gradually rotating the vectors according to the distance between each pixel and the boundary, as shown in Fig~\ref{intersec}. The rotation function is given by $R(\theta)$,  
\begin{equation}
\label{rotatefunction}
R(\theta) = \begin{bmatrix}
    \cos{\theta} & -\sin{\theta} \\
    \sin{\theta} & \cos{\theta} 
\end{bmatrix}
\end{equation}
where $\theta$ is defined by Eqt~\ref{theta}.
\begin{equation}
\label{theta}
\theta = \pi(1 - \mathbf{sigmoid}(S(p)))
\end{equation}
Putting Eqt~\ref{ac} and Eqt.~\ref{theta} together, we obtain Eqt~\ref{finaldynamicalsys}.
\begin{equation}
\label{finaldynamicalsys}
\frac{dp}{dt} = R(\theta) \nabla_p{S(p)}, 
\end{equation}
Eqt.~\ref{finaldynamicalsys} has an important property: When $p \in \partial{\Omega}$, $S(p) = 0$ so that $\theta$ is equal to $\frac{\pi}{2}$ according to Eqt.~\ref{rotatefunction}. This means on the boundary, the direction of $\frac{dp}{dt}$ is equal to the tangent of $p \in \partial{\Omega}$ as shown in step (d) in Fig~\ref{label_to_dynamic}. 

As opposed to active contour methods \cite{CootesActiveModels} where the dynamic is generated from images, we generate the discretized version of Eqt.~\ref{finaldynamicalsys} for each label. Then, a vector field is generated from it for each training instance with the property that limit cycle of the field is the boundary of ROI. This process generates a set of tuples (image, label, dynamic). That is, for each cardiac image, we have its associated binary label image, and its corresponding vector field. In the next subsection, we introduce the methodology to learn a CNN which maps an image patch to a vector from our vector field (Fig~\ref{pplaims}). This allows us to create an agent which follows step-by-step displacement predictions.

\subsection{Creating a Patch-Policy Predictor using a CNN}
\subsubsection{Training} 
\begin{figure}[h]
\begin{center}
  \includegraphics[scale=0.3]{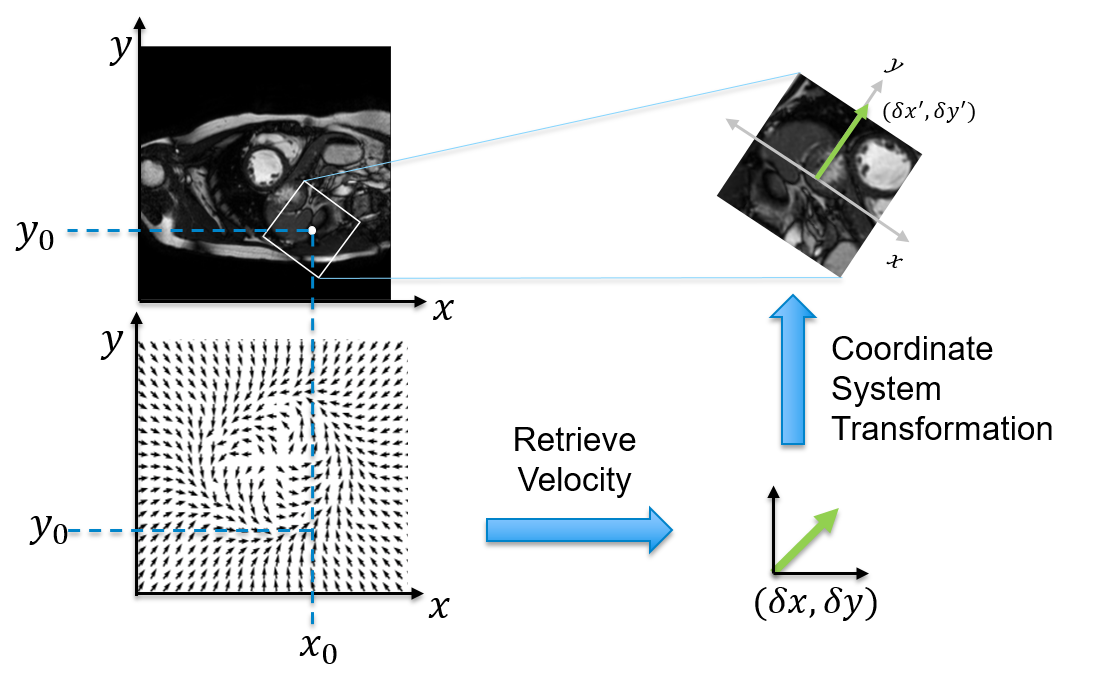}
  \caption{A patch extracted from the original image with its corresponding velocity in a vector field. The sampling patch's orientation is determined by the corresponding velocity. The velocity should be transformed into the coordinate system of the patch to be used as ground truth in training.}
  \label{sampledir}
  \end{center}
\vspace{-1.5em}  
\end{figure}

Our CNN operates over patches which are oriented with respect to our created dynamic. In order to prepare data for training, for each training image, we randomly choose a pre-defined proportion of points acting as the center of a rectangular sampling patch. We define a sampling direction which is equal to the velocity vector of the associated point. For example, for a given position $(x_0,y_0)$ on image, its velocity $(\delta x, \delta y)$ in the corresponding vector field is defined as the sampling direction, as shown in Fig~\ref{sampledir}. In the training process, such vectors are easily accessible, however they must be predicted during inference (see next Subsection~\ref{infer}). It is worth noting that a coordinate transformation is required to convert the velocity from the coordinate system of the dynamic to that of the patch, as illustrated in Fig~\ref{sampledir}. In order to improve robustness, training data augmentation can be performed by adding symmetric offsets to the sampling directions (e.g. (+\ang{45},-\ang{45})). Our CNN is based on the AlexNet architecture \cite{Krizhevsky2012ImageNetNetworks} with two output neurons. During training we use Adam optimizer with the mean square error (MSE) loss.\par 

\subsubsection{Inference}
\label{infer}
At the inference stage, before the first time step $t=0$, we determine an initial, rough, starting point using a basic LV detection module and a random sampling direction. This ensures that we don't start on an image boundary where there is insufficient input to create the first 64x64 pixel patch, and that we have an initial sampling direction. At each step, given an position $p_t$ and a sampling direction $s_t$ of the agent (which is unknown and is thus inferred as the difference between the current sampling direction and the last), a local patch is extracted and used as the input to the CNN-based policy model. The policy model then predicts the displacement for the agent to move, which in turn leads to the next local patch sample. This process iterates until the limit cycle is reached as illustrated earlier (Fig~\ref{wholeprocess_lite_1}).   

\subsection{Stopping Criterion: The Poincar\'e Map} 
\begin{figure}[h!]
  \begin{subfigure}[b]{0.35\textwidth}
    \includegraphics[width=\textwidth]{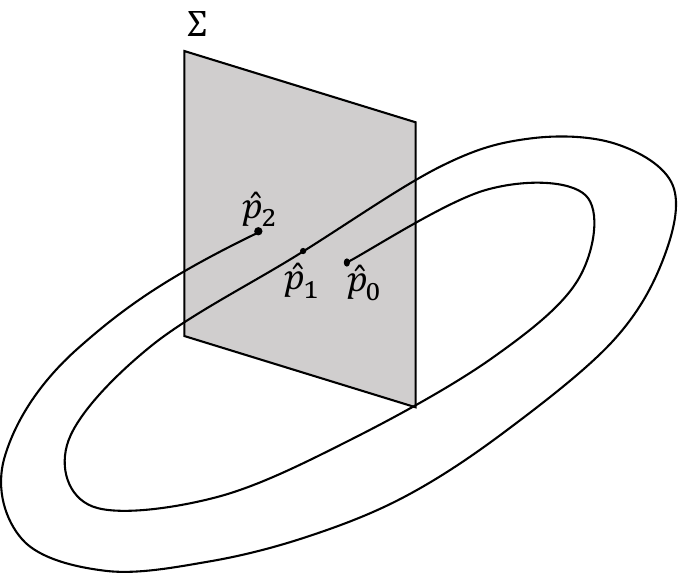}
    \caption{}
    \label{poincaresection_a}
  \end{subfigure}
  \hfill
  \begin{subfigure}[b]{0.35\textwidth}
    \includegraphics[width=\textwidth]{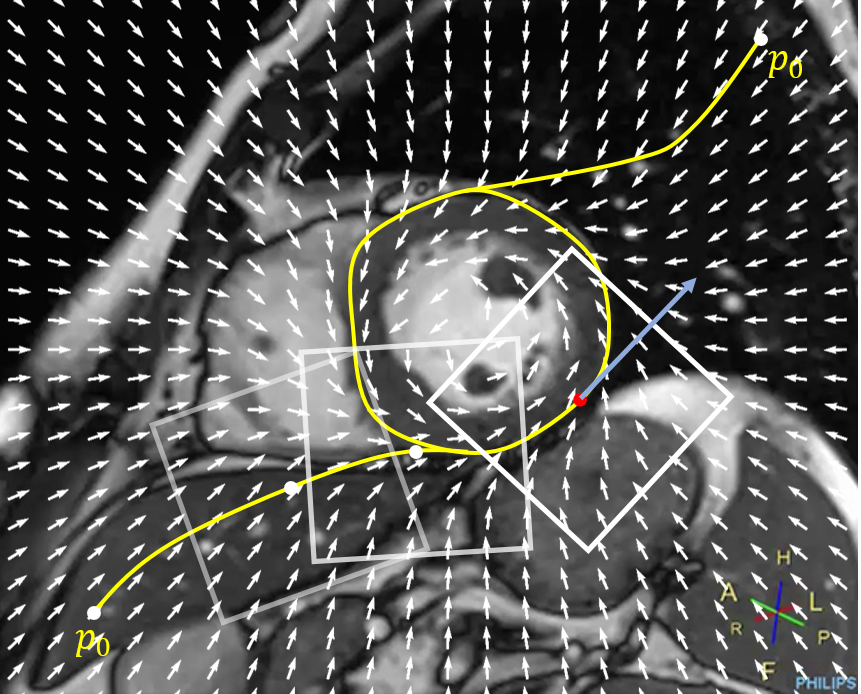}
    \caption{}
    \label{poincaresection_b}
  \end{subfigure}
  \caption{(a) An agent in a 3D space starting at $\hat{p}_0$ intersects the hyperplane (the Poincar\'e section) $\Sigma$ twice at $\hat{p}_1$ and $\hat{p}_2$. Performing analysis of the points on $\Sigma$ is much simpler and more efficient than the analysis of the trajectory in 3D space. (b) An agent starts at initial point $p_0$ on a cardiac MRI image. After $t$ iterations, the agent moves slowly toward the boundary of the object. Due to the underlying customized vector field, the DPM is able to guarantee that using different starting points we converge to the same unique periodic orbit (limit cycle).}
\vspace{-1.5em}  
\end{figure}

Instead of identifying the periodic orbit (the limit cycle) from the trajectory itself, we introduce the Poincar\'e section \cite{parker2012practical} which is a hyperplane, $\Sigma$, transversal to the trajectory. This cuts through the trajectory of the vector field, as seen in Fig~\ref{poincaresection_a}. The stability of a periodic orbit in the image can be reflected by the procession of corresponding points of intersection in $\Sigma$ (a lower dimensional space). The Poincar\'e map is the function which maps successive intersection points with the previous point, and thus, when the mapping reaches a small enough value we may say that the procession of the agent in the image has converged to the boundary (the limit cycle). The convergence of customized dynamic has been studied using the Poincar\'e-Bendixson theorem \cite{Parker1989PracticalSystems}, however the details are beyond the scope of this paper.

\section{Experimental Setting and Results}
\label{sec:EAR}
In this study, we evaluate our method on (1) the Sunnybrook Cardiac Dataset (SCD) \cite{RadauP.LuY.ConnellyK.PaulG.DickA.J.2009EvaluationMRI}, which contains 45 cases and (2) the STACOM 2011 LV Segmentation Challenge, which contains 100 cases.\par
\paragraph{\textbf{SCD Dataset}} The DPM was trained on the given training subset. We applied our trained model to the validation and online subsets (800 images from 30 cases in total) to provide a fair comparison with previous research, and we present our findings in Table \ref{tab:scd1}. We report the dice score, average perpendicular distance (APD) (in millimeters) and `good' contour rate (Good) for both the endocardium (i) and epicardium (o). We obtained a mean Dice score of 0.94 with a mean sensitivity of 0.95 and a mean specificity of 1.00. \par
\paragraph{\textbf{Transferability to the STACOM2011 Dataset}} 
To demonstrate the strong transferability of our method we train on the training subset of the SCD dataset and test on the STACOM 2011 dataset. We performed myocardium segmentation by segmenting the endocardium and epicardium separately, using 100 randomly selected MRI images from 100 cases. We report the Dice index, sensitivity, specificity, positive and negative predictive values (PPV and NPV) in Table \ref{tab:scd2}. We obtained a mean Dice index of 0.74 with a mean sensitivity of 0.84 and a mean specificity of 0.99.
\vspace{-1em}
\begin{table}[!h]
\centering
\caption{Comparison of LV endocardium and epicardium segmentation performance between DPM and previous research using the Sunnybrook Cardiac Dataset. Number format: mean value (standard deviation).}
\label{tab:scd1}
\begin{tabularx}{1.0\textwidth}{|l|XXXXXX|}
\hline
Method & Dice(i) & Dice(o) & $APD(i)$ & $APD(o)$ & $Good(i)$ & $Good(o)$ \\ \hline
\textbf{DPM} & 0.92(0.02) & \textbf{0.95(0.02)} & \textbf{1.75(0.45)} & \textbf{1.78(0.45)}  &  \textbf{97.5} &  \textbf{97.7} \\
Av2016\cite{avendi2016combined}   & \textbf{0.94(0.02)} & - & 1.81(0.44)  & -  &  96.69(5.7) & - \\
Qs2014\cite{queiros2014fast}   & 0.90(0.05) & 0.94(0.02)  & 1.76(0.45)  & 1.80(0.41) & 92.70(9.5)  & 95.40(9.6)  \\
Ngo2013\cite{ngo2013left}   & 0.90(0.03) & -  &  2.08(0.40)  & - & 97.91(6.2)  & -\\ 
Hu2013\cite{hu2013hybrid} & 0.89(0.03) & 0.94(0.02)  & 2.24(0.40)  &2.19(0.49)  &  91.06(9.4) &  91.21(8.5) \\    
       \hline 
\end{tabularx}
\end{table}
\vspace{-10mm}
\begin{table}[!h]
\centering
\caption{Comparison of myocardium segmentation performance by training on SCD data and testing on the STACOM 2011 LVSC dataset. Number format: mean value (standard deviation).}
\label{tab:scd2}
\begin{tabularx}{1.0\textwidth}{|l|XXXXX|}
\hline

Method &Dice& Sens. & Spec. & PPV& NPV \\ \hline
\textbf{DPM}& \textbf{0.74(0.15)}&  \textbf{0.84(0.20)}  & 0.99(0.01)  & 0.67(0.21)  &  0.99(0.01) \\
Jolly2012\cite{jolly2011automatic}   &0.66(0.25)&  0.62(0.27) & 0.99(0.01)  &\textbf{0.75(0.23)}  & 0.99(0.01)  \\
Margeta2012\cite{margeta2011layered}  &0.51(0.25)&  0.69(0.31)  & 0.99(0.01)  & 0.47(0.21) & 0.99(0.01)    \\
\hline 
\end{tabularx}
\end{table}

\section{Conclusion}
\label{sec:conclusion}
In this paper we have presented the Deep Poincar\'e Map as a novel method for LV segmentation and demonstrate its promising performance. 
The developed DPM method is robust for medical images, which have limited spatial resolution, low SNR and indistinct object boundaries. By encoding prior knowledge of a ROI as a customized dynamic, fine grained learning is achieved resulting in a displacement policy model for iterative segmentation. This approach requires much less training data than traditional methods. The strong transferability and rotational invariance of the DPM can be also attributed to this patch-based policy learning strategy. These two advantages are crucial for clinical applications.
\section{Acknowledgement}
Yuanhan Mo is sponsored by Sultan Bin Khalifa International Thalassemia Award. Guang Yang is supported by the British Heart Foundation Project Grant (PG/16/78/32402). Jingqing Zhang is supported by LexisNexis HPCC Systems Academic Program. Thanks to TensorLayer Community.

\bibliographystyle{splncs}
\bibliography{miccai_ref}

\end{document}